\documentclass{article} 
\usepackage{iclr2023_conference_tinypaper,times}

\usepackage{amsmath,amsfonts,bm}

\def\eqref#1{equation~\ref{#1}}

\def\1{\bm{1}}

\DeclareMathAlphabet{\mathsfit}{\encodingdefault}{\sfdefault}{m}{sl}
\SetMathAlphabet{\mathsfit}{bold}{\encodingdefault}{\sfdefault}{bx}{n}

\usepackage{hyperref}
\usepackage{url}
\usepackage{multirow}
\usepackage{makecell}
\usepackage[pdftex]{graphicx}

\title{Fast Adversarial CNN-based Perturbation Attack on No-Reference Image Quality Metrics}

\author{Ekaterina Shumitskaya$^1$, Anastasia Antsiferova$^{2,3}$ \& Dmitriy Vatolin$^{1,2,3}$  \\
$^1$ Lomonosov Moscow State University, Moscow, Russian Federation\\
$^2$ ISP RAS Research Center for Trusted Artificial Intelligence, Moscow, Russian Federation\\
$^3$ MSU Institute for Artificial Intelligence, Moscow, Russian Federation \\
\texttt{\{ekaterina.shumitskaya, aantsiferova, dmitriy\}@graphics.cs.msu.ru} \\
}

\iclrfinalcopy 
\begin{document}

\maketitle

\begin{abstract}
Modern neural-network-based no-reference image- and video-quality metrics exhibit performance as high as full-reference metrics. These metrics are widely used to improve visual quality in computer vision methods and compare video processing methods. However, these metrics are not stable to traditional adversarial attacks, which can cause incorrect results. Our goal is to investigate the boundaries of no-reference metrics applicability, and in this paper, we propose a fast adversarial perturbation attack on no-reference quality metrics. The proposed attack (FACPA) can be exploited as a preprocessing step in real-time video processing and compression algorithms. This research can yield insights to further aid in designing of stable neural-network-based no-reference quality metrics.

\end{abstract}

\section{Introduction}

State-of-the-art no-reference (NR) metrics commonly use neural-network-based approaches and deliver more accuracy than traditional ones. Quality-metric attacks lead such metrics to increase the quality score without improving visual quality. The developers of image- or video-processing and compression algorithms can exploit that vulnerability. For example, the developers of libaom \citep{deng2020vmaf} exploited the vulnerability of VMAF \citep{vmaf2017} by implementing the VMAF-oriented tuning option in the encoder. Our research aims to investigate the possibility of injecting quality-metric attacks into real-time algorithms, which can cause incorrect results in benchmarks. 
In this paper, we propose a Fast Adversarial CNN-based Perturbation Attack (FACPA) on NR quality metrics which is much faster than previous iterative attack methods and more effective than universal perturbation methods. Using this method, the researchers can estimate the potential cheating gain of any NR metric used in benchmarks and video processing algorithms.

\section{Related Work}

Existing attack methods mainly focus on attacking image classification models 
\citep{DBLP:journals/corr/SzegedyZSBEGF13},\citep{DBLP:journals/corr/GoodfellowSS14}, \citep{DBLP:conf/iclr/KurakinGB17a}, 
\citep{carlini2017towards}, \citep{DBLP:conf/iclr/MadryMSTV18}. Several studies on adversarial attacks related to NR quality metrics have been published. \cite{sang4112969generation} proposed an iterative algorithm based on adaptive distortions to attack quality metrics. This algorithm is based on the MI-FGSM \citep{dong2018boosting} and 
controls distortions using the NR metric NIQE \citep{mittal2012making}. 
\cite{korhonen2022adversarial} proposed a method for the iterative attack of NR quality metrics that employs Sobel filter to generate distortions on object edges since it allows to improve the visual quality of attacked images. \cite{zhang2022perceptual} proposed an iterative attack to craft adversarial images using different full-reference metrics
to control visual distortions. However, all these attacks are iterative and time-consuming, so their injection into real-time algorithms is unlikely. \cite{Shumitskaya_2022_BMVC} proposed to craft a universal adversarial perturbation to attack a metric. This approach is much faster than iterative methods. However, this attack does not use information about the input and hence does not use the full potential of the attack.

\section{Proposed Method and Results}

The proposed attack can be formulated as follows:
\begin{equation}
\underset{f}{\mathrm{argmax}} \{ \frac{1}{N}\sum_{i=1}^N (M(x_i + f(x_i)) - M(x_i))) \}, \forall x_i : \lVert f(x_i) \rVert _{\infty} \leq \epsilon,
\end{equation}

where \textit{f} is a function we a searching for, \textit{M} is the target metric, $x_i$ are images from the training set, $N$ is the training set size. To approximate this function, we use U-Net encoder-decoder architecture with the tanh activation and scaling. We tried different approaches and chose U-Net architecture as it showed the best performance on preliminary tests. Scaling provides a constraint $ \epsilon $. We used $ \epsilon $ equal to $ \frac{10}{255} $. For each target metric, we trained specific CNN weights. For CNNs training, we used 10,000 256 $ \times $ 256 images from the COCO dataset \citep{lin2014microsoft}, defined the loss function as the target metric with the opposite sign and employed Adam optimizer \citep{kingma2014adam}.

\begin{figure}[h]
\begin{center}
\includegraphics[width=5.2in]{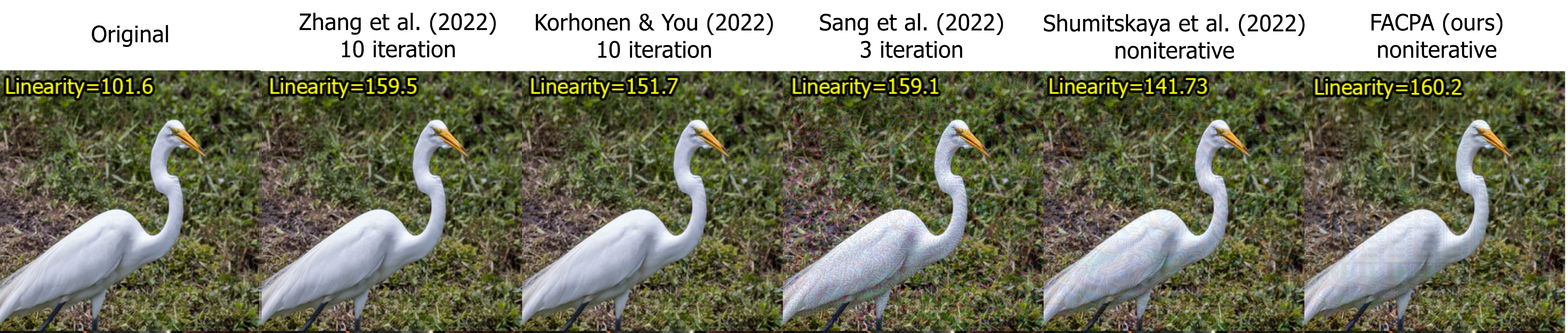}
\end{center}
\caption{Attack examples for the Linearity quality metric.}
\label{attacked_images}
\end{figure}

We compared the proposed attack with previous attacks on quality metrics using images from NIPS 2017: Adversarial Learning Development Set \citep{nips2017}. We attacked three quality metrics (Linearity \citep{li2020norm}, VSFA \citep{li2019quality} and MDTVSFA \citep{li2021unified}) and compared the achieved increase of target metrics scores. For iterative attacks, we used a learning rate of 0.001
and attacked metrics using a different amount of
iterations. Table \ref{results} contains comparison results.
FACPA is much faster than iterative methods (15 ms versus 110 -- 5,000 ms) and more effective than the universal perturbation method. Figure \ref{attacked_images} shows attack examples for the Linearity metric for compared methods.

\begin{table}[h]
\caption{Percentage of metrics increase by proposed and other attacks and GPU calculation time.}
\label{results}
\begin{center}
\begin{tabular}{cccccccc}
\multirow{2}{*}{\textbf{Attack method}} & \multirow{2}{*}{\makecell{It.}} & \multicolumn{2}{c}{Linearity} & \multicolumn{2}{c}{VSFA} & \multicolumn{2}{c}{MDTVSFA} \\ 
     & & \makecell{Gain $\uparrow$} & Speed $\downarrow$ & \makecell{Gain $\uparrow$} & Speed $\downarrow$ & \makecell{Gain $\uparrow$} & Speed $\downarrow$ \\ 
     \hline
     \multirow{2}{*}{\makecell{\cite{zhang2022perceptual}}} & 1 & 5.9\% & 173 ms & 11.7\% & 626 ms & 21.3\% & 631 ms \\ 
      & 10 & \underline{26.1}\% & 1590 ms & 30.9\% & 2660 ms & \textbf{47.9}\% & 5980 ms \\ 
     \multirow{2}{*}{\makecell{\cite{korhonen2022adversarial}}} & 1 & 5.6\% & 111 ms & 11.1\% & 483 ms & 20.4\% & 505 ms \\
      & 10 & 21.0\% & 992 ms & \textbf{41.7}\% & 4750 ms & 39.3\% & 4040 ms \\
     \multirow{2}{*}{\makecell{\cite{sang4112969generation}}} & 1 & 6.5\% & 153 ms & 4.4\% & 535 ms & 3.7\% & 540 ms \\
      & 3 & 18.6\% & 358 ms & 32.6\% & 1580 ms & 39.2\% & 1580 ms \\
     \makecell{\cite{Shumitskaya_2022_BMVC}} & - & 22.9\% & \textbf{11 ms} & 22.0\% & \textbf{11 ms} & 29.4\% & \textbf{11 ms} \\
     \makecell{FACPA (ours) } & - & \textbf{32.0}\% & \underline{15 ms} & 30.6\% & \underline{15 ms} & 40.3\% & \underline{15 ms} \\
\end{tabular}
\end{center}
\end{table}

\section{Conclusion}

In this paper, we proposed a fast CNN-based attack that increases the scores of NR quality metrics. The comparison with previous methods showed that our attack is much faster than iterative methods and more efficient than the universal adversarial perturbation. The speed of the proposed
attack shows that it can be injected in video compression and other real-time algorithms. Therefore proposed attack can serve as an additional verification of metric reliability. Our code is publicly available at \url{https://github.com/katiashh/FACPA}.





\subsubsection*{URM Statement}
The authors acknowledge that at least one key author of this work meets the URM criteria of ICLR 2023 Tiny Papers Track.

\subsubsection*{Acknowledgements}
The work received support through a grant for research centers in the field of artificial intelligence
(agreement identifier 000000D730321P5Q0002, dated November 2, 2021, no. 70-2021-00142 with
the Ivannikov Institute for System Programming of the Russian Academy of Sciences).

\bibliography{egbib}

\begin{thebibliography}{19}
\providecommand{\natexlab}[1]{#1}
\providecommand{\url}[1]{\texttt{#1}}
\expandafter\ifx\csname urlstyle\endcsname\relax
  \providecommand{\doi}[1]{doi: #1}\else
  \providecommand{\doi}{doi: \begingroup \urlstyle{rm}\Url}\fi

\bibitem[Alex~K(2017)]{nips2017}
Ian~Goodfellow Alex~K, Ben~Hamner.
\newblock Nips 2017: Defense against adversarial attack, 2017.
\newblock URL
  \url{https://kaggle.com/competitions/nips-2017-defense-against-adversarial-attack}.

\bibitem[Carlini \& Wagner(2017)Carlini and Wagner]{carlini2017towards}
Nicholas Carlini and David Wagner.
\newblock Towards evaluating the robustness of neural networks.
\newblock In \emph{2017 ieee symposium on security and privacy (sp)}, pp.\
  39--57. Ieee, 2017.

\bibitem[Deng et~al.(2020)Deng, Han, and Xu]{deng2020vmaf}
Sai Deng, Jingning Han, and Yaowu Xu.
\newblock Vmaf based rate-distortion optimization for video coding.
\newblock In \emph{2020 IEEE 22nd International Workshop on Multimedia Signal
  Processing (MMSP)}, pp.\  1--6. IEEE, 2020.

\bibitem[Dong et~al.(2018)Dong, Liao, Pang, Su, Zhu, Hu, and
  Li]{dong2018boosting}
Yinpeng Dong, Fangzhou Liao, Tianyu Pang, Hang Su, Jun Zhu, Xiaolin Hu, and
  Jianguo Li.
\newblock Boosting adversarial attacks with momentum.
\newblock In \emph{Proceedings of the IEEE conference on computer vision and
  pattern recognition}, pp.\  9185--9193, 2018.

\bibitem[Goodfellow et~al.(2015)Goodfellow, Shlens, and
  Szegedy]{DBLP:journals/corr/GoodfellowSS14}
Ian~J. Goodfellow, Jonathon Shlens, and Christian Szegedy.
\newblock Explaining and harnessing adversarial examples.
\newblock In Yoshua Bengio and Yann LeCun (eds.), \emph{3rd International
  Conference on Learning Representations, {ICLR} 2015, San Diego, CA, USA, May
  7-9, 2015, Conference Track Proceedings}, 2015.
\newblock URL \url{http://arxiv.org/abs/1412.6572}.

\bibitem[Kingma \& Ba(2014)Kingma and Ba]{kingma2014adam}
Diederik~P Kingma and Jimmy Ba.
\newblock Adam: A method for stochastic optimization.
\newblock \emph{arXiv preprint arXiv:1412.6980}, 2014.

\bibitem[Korhonen \& You(2022)Korhonen and You]{korhonen2022adversarial}
Jari Korhonen and Junyong You.
\newblock Adversarial attacks against blind image quality assessment models.
\newblock In \emph{Proceedings of the 2nd Workshop on Quality of Experience in
  Visual Multimedia Applications}, pp.\  3--11, 2022.

\bibitem[Kurakin et~al.(2017)Kurakin, Goodfellow, and
  Bengio]{DBLP:conf/iclr/KurakinGB17a}
Alexey Kurakin, Ian~J. Goodfellow, and Samy Bengio.
\newblock Adversarial examples in the physical world.
\newblock In \emph{5th International Conference on Learning Representations,
  {ICLR} 2017, Toulon, France, April 24-26, 2017, Workshop Track Proceedings}.
  OpenReview.net, 2017.
\newblock URL \url{https://openreview.net/forum?id=HJGU3Rodl}.

\bibitem[Li et~al.(2019)Li, Jiang, and Jiang]{li2019quality}
Dingquan Li, Tingting Jiang, and Ming Jiang.
\newblock Quality assessment of in-the-wild videos.
\newblock In \emph{Proceedings of the 27th ACM International Conference on
  Multimedia}, pp.\  2351--2359, 2019.

\bibitem[Li et~al.(2020)Li, Jiang, and Jiang]{li2020norm}
Dingquan Li, Tingting Jiang, and Ming Jiang.
\newblock Norm-in-norm loss with faster convergence and better performance for
  image quality assessment.
\newblock In \emph{Proceedings of the 28th ACM International Conference on
  Multimedia}, pp.\  789--797, 2020.

\bibitem[Li et~al.(2021)Li, Jiang, and Jiang]{li2021unified}
Dingquan Li, Tingting Jiang, and Ming Jiang.
\newblock Unified quality assessment of in-the-wild videos with mixed datasets
  training.
\newblock \emph{International Journal of Computer Vision}, 129\penalty0
  (4):\penalty0 1238--1257, 2021.

\bibitem[Lin et~al.(2014)Lin, Maire, Belongie, Hays, Perona, Ramanan,
  Doll{\'a}r, and Zitnick]{lin2014microsoft}
Tsung-Yi Lin, Michael Maire, Serge Belongie, James Hays, Pietro Perona, Deva
  Ramanan, Piotr Doll{\'a}r, and C~Lawrence Zitnick.
\newblock Microsoft coco: Common objects in context.
\newblock In \emph{European conference on computer vision}, pp.\  740--755.
  Springer, 2014.

\bibitem[Madry et~al.(2018)Madry, Makelov, Schmidt, Tsipras, and
  Vladu]{DBLP:conf/iclr/MadryMSTV18}
Aleksander Madry, Aleksandar Makelov, Ludwig Schmidt, Dimitris Tsipras, and
  Adrian Vladu.
\newblock Towards deep learning models resistant to adversarial attacks.
\newblock In \emph{6th International Conference on Learning Representations,
  {ICLR} 2018, Vancouver, BC, Canada, April 30 - May 3, 2018, Conference Track
  Proceedings}. OpenReview.net, 2018.
\newblock URL \url{https://openreview.net/forum?id=rJzIBfZAb}.

\bibitem[Mittal et~al.(2012)Mittal, Soundararajan, and Bovik]{mittal2012making}
Anish Mittal, Rajiv Soundararajan, and Alan~C Bovik.
\newblock Making a “completely blind” image quality analyzer.
\newblock \emph{IEEE Signal processing letters}, 20\penalty0 (3):\penalty0
  209--212, 2012.

\bibitem[Sang et~al.(2022)Sang, Zhang, Liu, Wu, and
  Bovik]{sang4112969generation}
Qingbing Sang, Hongguo Zhang, Lixiong Liu, Xiaojun Wu, and Alan Bovik.
\newblock On the generation of adversarial samples for image quality
  assessment.
\newblock \emph{Available at SSRN 4112969}, 2022.

\bibitem[Shumitskaya et~al.(2022)Shumitskaya, Antsiferova, and
  Vatolin]{Shumitskaya_2022_BMVC}
Ekaterina Shumitskaya, Anastasia Antsiferova, and Dmitriy~S Vatolin.
\newblock Universal perturbation attack on differentiable no-reference image-
  and video-quality metrics.
\newblock In \emph{33rd British Machine Vision Conference 2022, {BMVC} 2022,
  London, UK, November 21-24, 2022}. {BMVA} Press, 2022.
\newblock URL \url{https://bmvc2022.mpi-inf.mpg.de/0790.pdf}.

\bibitem[Szegedy et~al.(2014)Szegedy, Zaremba, Sutskever, Bruna, Erhan,
  Goodfellow, and Fergus]{DBLP:journals/corr/SzegedyZSBEGF13}
Christian Szegedy, Wojciech Zaremba, Ilya Sutskever, Joan Bruna, Dumitru Erhan,
  Ian~J. Goodfellow, and Rob Fergus.
\newblock Intriguing properties of neural networks.
\newblock In Yoshua Bengio and Yann LeCun (eds.), \emph{2nd International
  Conference on Learning Representations, {ICLR} 2014, Banff, AB, Canada, April
  14-16, 2014, Conference Track Proceedings}, 2014.
\newblock URL \url{http://arxiv.org/abs/1312.6199}.

\bibitem[VMAF: Perceptual video quality assessment based on multi-method
  fusion()]{vmaf2017}
VMAF: Perceptual video quality assessment based on multi-method fusion.
\newblock Vmaf: Perceptual video quality assessment based on multi-method
  fusion, 2017.
\newblock URL \url{https://github.com/Netflix/vmaf}.

\bibitem[Zhang et~al.(2022)Zhang, Li, Min, Zhai, Guo, Yang, and
  Ma]{zhang2022perceptual}
Weixia Zhang, Dingquan Li, Xiongkuo Min, Guangtao Zhai, Guodong Guo, Xiaokang
  Yang, and Kede Ma.
\newblock Perceptual attacks of no-reference image quality models with
  human-in-the-loop.
\newblock \emph{arXiv preprint arXiv:2210.00933}, 2022.

\end{thebibliography}
\bibliographystyle{iclr2023_conference_tinypaper}


\end{document}